\documentclass[
]{ceurart}

\usepackage{listings}
\begin{document}

\copyrightyear{2022}
\copyrightclause{Copyright for this paper by its authors.
  Use permitted under Creative Commons License Attribution 4.0
  International (CC BY 4.0).}
\title{CERIST’22: Classifying COVID-19 Related Tweets for Fake News Detection and Sentiment Analysis with BERT-based Models}

\conference{CERIST'22: CERIST NLP Challenge 2022,
  March 29, 2023, Algeria, Algiers}
.
\author[1]{Rabia Bounaama}[%
email=rabea.bounaama@univ-tlemcen.dz, 
]
\cormark[1]
\fnmark[1]
\address[1]{Biomedical Engineering Laboratory,Tlemcen University,Algeria,}

\author[2]{Mohammed El Amine Abderrahim}[%
email=mohammedelamine.abderrahim@univ-tlemcen.dz, 
]
\fnmark[1]
\address[2]{Laboratory of Arabic Natural Language Processing,Tlemcen University,Algeria,}

\cortext[1]{Corresponding author.}

\begin{abstract}
The present paper is about the participation of our team “techno” on CERIST’22 shared tasks. We used an available dataset “task1.c” related to covid-19 pandemic. It comprises 4128 tweets for sentiment analysis task and 8661 tweets for fake news detection task. We used natural language processing tools with the combination of the most renowned pre-trained language models BERT (Bidirectional Encoder Representations from Transformers). The results shows the efficacy of pre-trained language models as we attained an accuracy of 0.93 for the sentiment analysis task and 0.90 for the fake news detection task.
\end{abstract}


\maketitle

\section{Introduction}

Covid-19 has affected each individual’s daily life, on October 2022, over of 600 million cases have been reported around the world\footnote{https://www.worldometers.info/coronavirus/}. Under the current disruption caused by covid-19, everyone is responsible to government measures aimed at preventing the spread of the virus. Includes distance learning, quarantine, staying at home and social distancing. Those decisions have results an amount of user’s opinion on social media. One of the most effective methods for analyzing these opinions is Sentiment Analysis (SA). SA is a classification process that identifies the opinions and emotions of users through the written contents \cite{hassonah2020efficient}. 

Information on social media platforms may contain true and false facts. Unfortunately, false facts spread rapidly \cite{stahl2018fake}. Fake news (FN) is the deliberate spread of misinformation via traditional news media or via social media  \cite{stahl2018fake}.The critical situation imposed by the covid-19 pandemic has increased people’s attention particularly regards to the number of new cases and deaths. If this information is wrong, it can have a negative impact on people's mental health, like depression and anxiety disorder. For that reason FN detection has become an urgent area of research for researcher community. FN detecting is to predict whether the news article is a FN piece or not \cite{shu2017fake}. In this context, there is pressing need for practical studies. To this end CERIST has proposed several tasks to solve. Our participation is related to opinion mining and SA task, precisely “task1.c: Arabic sentiment analysis and fake news detection within COVID-19”\footnote{http://www.nlpchallenge.cerist.dz}.

To address SA and FN detection tasks several approaches can be used. Among the most popular that has shown better results is the transfer learning approach using BERT. The use of pre-trained BERT models allows us to fine-tune the model for the specific task without having to train the model from scratch. This involves training the model on a labeled dataset provided by the challenge organizers. The model's biases and weights were adjusted to fit the requirements of either the SA or FN detection task. After training, the model was used to predict the labels on an unseen test dataset.\\

The remainder of this paper is organized as follow. We present in Section 2 related work on SA and FN detection in comments about covid-19. Section 3 describes the experiment. The obtained results are discussed in section 4. Finally, section 5 concludes the paper.

\section{Related work}
This related work is divided into two sections: Arabic SA related covid-19 and Arabic FN detection related covid-19.

Several studies have been conducted to address SA related to Covid-19 pandemic and different approaches and methods are applied. Among these approaches, deep learning is a commonly used tool especially for short texts such as tweets. The study by  \cite{alhumoud2020arabic}on Arabic SA applied deep learning to a corpus of 416,292 annotated tweets about covid-19. Authors found that the word ‘corona’ and ‘virus’ are dominant in their corpus. They used traditional Machine Learning (ML) such as Support Vector Machines (SVM), the new pre-trained transformer model “arabert” and ensemble model. As results they found that negative sentiment is dominant as opposed to the positive and neutral.  They conclude that the proposed ensemble model outperforms SVM, deep learning models and AraBERT with 0.90 as accuracy.

Covid-19 had an important impact on government decision in all countries worldwide. For example, Saudi Arabia implemented a series of control measures on order to deal with the pandemic. The two following studies concern the SA of tweets in Saudi Arabia regarding distance learning during Covid-19 \cite{aljabri2021sentiment}, and regarding governmental preventive measure to contain Covid-19\cite{alhajji2020sentiment}. In the study of \cite{alhajji2020sentiment}, authors used Naive Bayes (NB) as ML model to train a corpus of 53,127 tweets containing hashtags related to seven public health measures imposed by the Saudi government. The measures are: Grand Mosque closure, Qatif lockdown, school and university closure, shopping malls, parks and restaurants closure, sports competition suspension, congregational and weekly Friday prayers suspension and nationwide curfew. The results of their analysis conclude that twitter users in Saudi Arabia support the applicable government measure except one measure: shopping malls, parks, and restaurant closure. For the six other measures, results show that positive sentiment are more than negative ones especially for measure that pertain religious practice.

The research conducted by \cite{aljabri2021sentiment}showed the acceptance rate of the distance learning enforcement in Saudi Arabia. Their sentiment classifier was built using well-known ML and feature extraction techniques. As best result, Logistic Regression (LR) with the use of unigrams and tf-idf approaches for feature extraction had an accuracy of 0.89. Their research concludes that positive opinions are more numerous than negative ones. 

The work described in Manal M.A., et al  \cite{ali2021arabic} concerns e-learning to mitigate Covid-19. These authors also used traditional ML like SVM, K Nearest Neighbor (KNN), Multinomial Naïve Bayes (MNB) and LR. The data extracted from twitter, via twitter APIs, are preprocessed and labeled depending on the emotional weights of eight emotions using word-emotion lexicon NRC (the National Research Council Canada). As best result, SVM classifier outperforms the others classifiers with 0.89 as accuracy.

For the Arabic language, there are few studies and datasets published during Covid-19 about FN detection and it hasn't achieved the level compared to other languages, especially English \cite{mahlous2021fake}. In the following section, we present some recent automated systems for Arabic FN detection during Covid-19.
 The study of Ahmed .R.M et al  \cite{mahlous2021fake} introduces a public dataset within Covid-19 FN detection in Arabic tweets. The authors compare two built corpus: 2500 manually labeled tweets and 34529 automated labeled tweets from 7 million extracted tweets. By using classical ML, The results show that the automatically and manually labeled corpora reached 0.93 and 0.87 respectively as f1 score.
        
In  \cite{shishah2022jointbert}, authors tested transformer models with a new strategy “JointBERT” in four FN dataset: Covid19Fakes, AraNews, Satirical News and ANS. They proposed a new technique in detecting Arabic FN. The proposed technique had an improvement reach to 10\% comparing to baseline Arabic FN detection models. Authors used their proposed method “JointBERT” for sequence labeling and encoding process, more details about their proposed model are in\cite{shishah2021fake}.

In\cite{nassif2022arabic}, authors constructed two Arabic FN detection datasets related to Covid-19. The first dataset comprises 16,000 examples and was translated to English from “Kaggle”, while the second dataset comprises 10,000 examples and was collected from Arabic websites and tweets. The authors developed eight transformer models, including Giga-Bert, Roberta-Base, AraBert, Arabic-BERT, ARBERT, MarBert, Araelectra and QaribBert, as well as contextualized Arabic embedding models. The developed models performed much better on the Arabic dataset compared to the translated one. ARBERT pre-trained model achieved 0.98 as accuracy. 

\section{Experiments}
In this section we describe the followed steps for building SA model, and FN detection model.

\subsection{Pre-trained models description }

Transformer is a simple network architecture proposed by \cite{vaswani2017attention}, which is based on the attention mechanism \cite{bahdanau2014neural}. The main idea behind the attention mechanism is to allow the encoder-decoder model to automatically search for relevant parts from the input source in order to predict a relevant target word for machine translation purposes by extending the vector length \cite{cho2014learning}.

BERT \cite{devlin2018bert}is a language representation. For text classification BERT-base model contains an encoder with 12 Transformer blocks. BERT takes an input of a sequence of no more than 512 tokens and outputs the representation of sequence  \cite{sun2019fine}. The pre-trained BERT model becomes a state-of-the-art of several NLP tasks such as question answering and language inference \cite{devlin2018bert}.

We implemented the pre-trained bert-mini-arabic and bert-base-arabic\cite{safaya2020kuisail} for SA task and FN detection task respectively from hungging face library\footnote{https://huggingface.co/}. The model was pre-trained on approximately 8.2 Billion words. The model is a combination of convolutional neural network (CNN) with BERT \cite{safaya2020kuisail}.
\subsection{Data preprocessing}
We perform data distribution before preprocessing it. As shown in the table below (see table\ref{tab:lab1}), the training data for the SA task and the FN detection task has a balanced distribution of classes, which may impact the model’s accuracy performance.

We implement some known setups for preparing the training dataset for BERT models. Knowing that the BERT model relies heavily on the context of a word in a sentence, then, during data cleaning we try to preserve the semantic and syntactic structure of each sentence “tweet”.

We used regular expressions to remove URLs and punctuations marks by using the python “String” module. We replaced certain punctuation marks in the text with space and we removed emoji symbols. We kept digits unchanged. Note that we did not find any duplicate tweets and the established pre-process was implemented for both datasets. Finally data was ready for training stage using BERT pre-trained models.

\begin{table*} [!ht]
  \caption{Datasets  distribution }
  \label{tab:lab1}
  \begin{tabular}{ccl}
    \toprule
   Task & Count values per each class\\
    \midrule 
    SA (sentiment analysis) & neutral sentiment  1600\\
    & positive sentiment 1571\\
    &negative sentiment  957\\
    \midrule 
    Total &4128 tweets\\
    \midrule 
    FN (fake news) & Can’t decide 5024\\
    &maybe  1471\\
    &yes 367\\
    &no 1799\\
    \midrule 
Total &8661tweets\\
  \bottomrule
\end{tabular}
\end{table*}
As indicated in Table 1, the corpus of total tweets in the SA task consists of 38\% neutral and positive sentiments. In the FN task, the "can't decide" class accounts for 58\%, while the "yes" and "no" classes make up 4\% and 16\% of the corpus, respectively. Predicting minority classes may be difficult due to the impact of imbalanced datasets on most ML algorithms \cite{zhang2020revisiting}.
\subsection{Experimental fine-tuning }
The fine-tuning process is an essential step in using pre-trained models like BERT. It involves adjusting the parameters of the pre-trained model to suit the specific task that it's being applied to. For the BERT models, in our case, we used the following fine-tuning parameters:
\begin{itemize}
    \item Training batch size: We used a batch size of 10 during the training process. The batch size is the number of samples that are processed at once before the model's weights are updated.
\item Learning Learning rate: The learning rate determines the step size at which the optimizer updates the model's weights. For our models, we used a learning rate of 2e-5.
    \item Optimizer: AdamW is the optimizer that we used to update the model's weights during the training process. AdamW is a variation of the Adam optimizer that's well-suited for fine-tuning pre-trained models like BERT.
     \item Train epochs: The number of training epochs refers to the number of times that the model is trained on the entire dataset. For our models, we used a total of 10 training epochs.
    \item Maximum sequence length: The maximum sequence length is the maximum number of tokens that can be fed into the BERT model at once. In our case, we used a maximum sequence length of 256.

\end{itemize}
By using these fine-tuning parameters, we were able to effectively fine-tune the pre-trained BERT models for our SA and FN detection tasks. 

\subsection{Results and discussion }

To evaluate the performance of our models, we used the fine-tuned BERT models to predict the targets for both the SA task and FN detection task and submitted our results to the challenge committee.

 In the absence of the results obtained by the other participants in the challenge, we cannot make a comparison with our results. Nevertheless, we can say that the use of the pre-trained language models BERT gives good results in the context of the SA and the FN detection.

As shown in the tables below (Table \ref{tab:lab2} and Table \ref{tab:lab3}),the SA task result, in term of the three performance measures, outperforms the FN detection task. This result can be attributed to the balanced distribution of classes in the SA task dataset.

The FN detection dataset is unbalanced, which could lead to bias in the output and affect the overall performance of the model.

\begin{table*}[!ht]
  \caption{Techno-team results for task1.c on validation dataset for sentiment analysis and fake news detection tasks   }
  \label{tab:lab2}
  \begin{tabular}{cclll}
    \toprule
  \\ Task & precision & recall & f1-score\\
    \midrule 
    SA & 0.94 &     0.93  &    0.94 \\
    \midrule 
   FN &  0.87 &     0.86     & 0.86\\
 
  \bottomrule
\end{tabular}
\end{table*}
\begin{table*} [!ht]
  \caption{Techno-team results for task1.c on test dataset for sentiment analysis and fake news detection tasks   }
  \label{tab:lab3}
  \begin{tabular}{cclll}
    \toprule
  \\ Task & accuracy & f1-score\\
    \midrule 
    SA (run1) & 0.9332 &     0.9332   \\
    SA (run2) & 0.9332 &     0.9331  \\
    SA (run3) & 0.9381 &     0.9381 \\
    \midrule 
   FN &  0.9021 &     0.9016   \\
 
  \bottomrule
\end{tabular}
\end{table*}

 \section{Conclusion }
 
In this work, we utilized natural language processing and pre-trained language models BERT to address two classification subtasks as part of CERIST’22 shared tasks “task1.c”. We obtained  0.93 for f1-score on sentiment analysis task and a 0.90 for fake news detection task. These results indicated that this approach effectively solved the sentiment analysis task related to COVID-19. However, the limitations of the model are evident in the unbalanced dataset of the fake news detection task. To further improve our results, in future work, we plan to implement other learning approach.   
 
 \bibliography{sample-ceur}
\end{document}